\documentclass[10pt,twocolumn,letterpaper]{article}

\usepackage{cvpr}
\cvprfinalcopy
\usepackage{times}
\usepackage{epsfig}
\usepackage{graphicx}
\usepackage{amsmath}
\usepackage{amssymb}
\usepackage{cite}
\usepackage{lib}
\usepackage{alias}
\usepackage{booktabs}
\usepackage{multirow}
\usepackage[skip=4pt,font=small,labelfont=bf]{caption}
\usepackage[pagebackref=true,breaklinks=true,letterpaper=true,colorlinks,bookmarks=false,citecolor=blue]{hyperref}

\cvprfinalcopy 

\def\cvprPaperID{6583} 

\def\methodname{Neural Topological SLAM}
\def\methodabbr{NTS}

\makeatletter
\renewcommand*{\@fnsymbol}[1]{\ensuremath{\ifcase#1\or \dagger\or *\or \ddagger\or
    \mathsection\or \mathparagraph\or \|\or **\or \dagger\dagger
    \or \ddagger\ddagger \else\@ctrerr\fi}}
\makeatother

\ifcvprfinal\fi
\begin{document}

\title{Neural Topological SLAM for Visual Navigation}

\author{
Devendra Singh Chaplot\textsuperscript{\textnormal{1}}\thanks{Correspondence: \texttt{chaplot@cs.cmu.edu}}  , Ruslan Salakhutdinov\textsuperscript{\textnormal{1}}\footnotemark[2] , Abhinav Gupta\textsuperscript{\textnormal{1,2}}\footnotemark[2] , Saurabh Gupta\textsuperscript{\textnormal{3}}\thanks{Equal Contribution} \\
\textsuperscript{1} Carnegie Mellon University, \textsuperscript{2} Facebook AI Research, \textsuperscript{3} UIUC \\[10pt]
\small{Project webpage: \url{https://devendrachaplot.github.io/projects/Neural-Topological-SLAM}}
}

\maketitle


\begin{abstract}
\vspace{-6pt}
This paper studies the problem of image-goal navigation which involves navigating to the location indicated by a goal image in a novel previously unseen environment. To tackle this problem, we design topological representations for space that effectively leverage semantics and afford approximate geometric reasoning. At the heart of our representations are nodes with associated semantic features, that are interconnected using coarse geometric information. We describe supervised learning-based algorithms that can build, maintain and use such representations under noisy actuation. Experimental study in visually and physically realistic simulation suggests that our method builds effective representations that capture structural regularities and efficiently solve long-horizon navigation problems. We observe a relative improvement of more than 50\% over existing methods that study this task.
\vspace{-6pt}
\end{abstract}

\section{Introduction}
\begin{figure}
\centering
\includegraphics[width=\linewidth,height=\textheight,keepaspectratio]{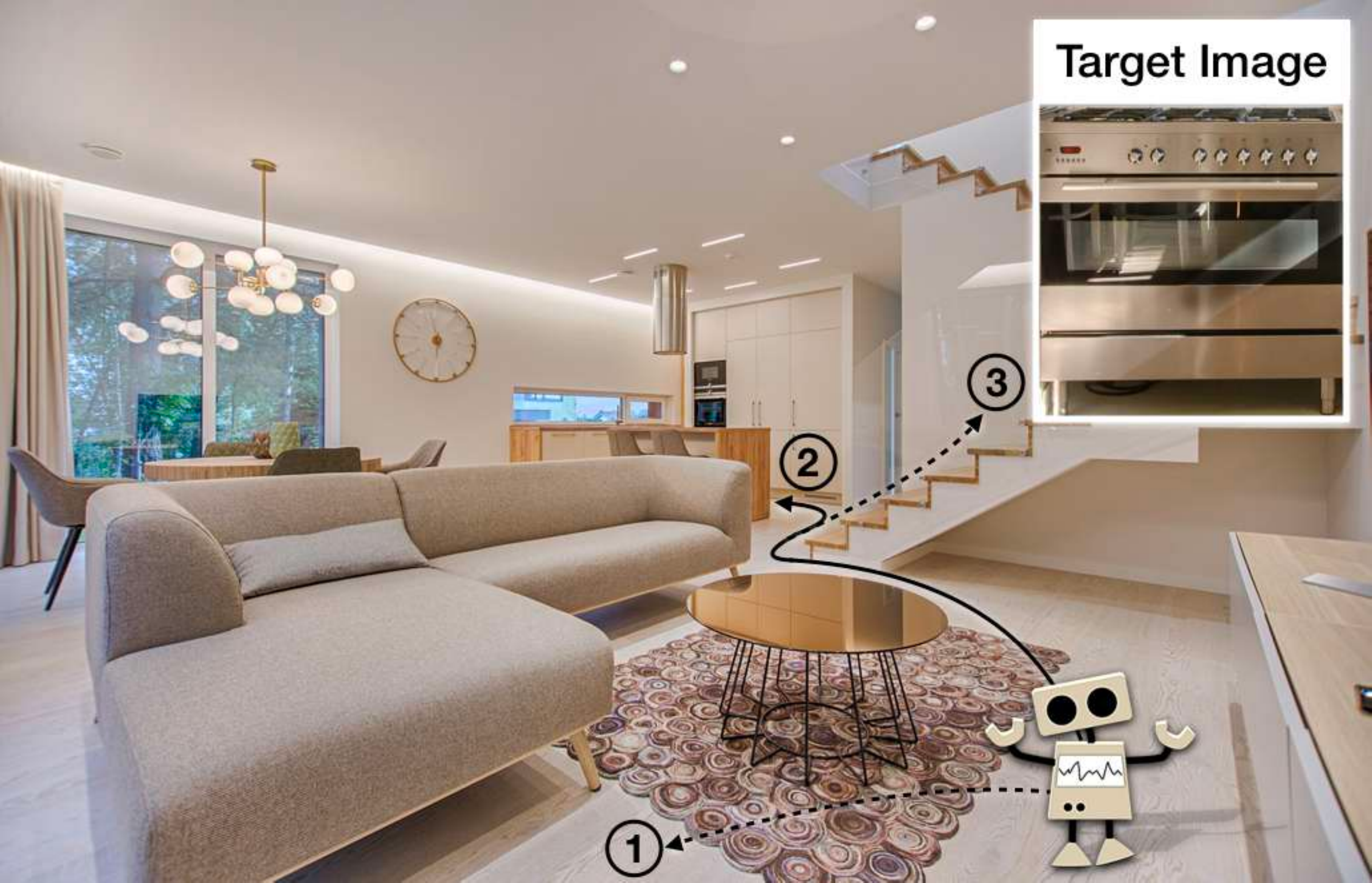}
\vspace{-12pt}
\caption{\small \textbf{Semantic Priors and Landmarks.} When asked to go to target image of an oven most humans would use the path number 2 since it allows access to kitchen. Humans use semantic priors and common-sense to explore and navigate everyday yet most navigation algorithms struggle to do so.}
\label{fig:teaser}
\vspace{-5pt}
\end{figure}

Imagine you are in a new house as shown in Fig~\ref{fig:teaser} and you are given the task of finding a target object as shown in Fig~\ref{fig:teaser} (top). While there are multiple possible directions to move, most of us would choose the path number 2 to move. This is because we use strong structural priors -- we realize the target is an oven which is more likely to be found in the kitchen which seems accessible via path number 2. Now let us suppose, once you reach the oven, your goal is to reach back to the living room which you saw initially. How would you navigate?  The answer to this question lies in how we humans store maps (or layout) of the house we just traversed. One possible answer would be metric maps, in which case we would know exactly how many steps to take to reach the living room. But this is clearly not how we humans operate~\cite{gillner1998navigation,wang2002human}. Instead, most of us would first get out of the kitchen by moving to the hallway and then navigate to the living room which is visible from the hallway. 

It is clear from the above examples, there are two main components of a successful visual navigation algorithm: (a) ability to build spatial representations and store them; (b) ability to exploit structural priors. When it comes to spatial representations, the majority of papers in navigation insist on building metrically precise representations of free space. However, metric maps have two major shortcomings: first, metric maps do not scale well with environment size and amount of experience. But more importantly, actuation noise on real-robots makes it challenging to build consistent representations, and precise localization may not always be possible. When it comes to exploiting structural priors, most learning-based approaches do not model these explicitly. Instead, they hope the learned policy function has these priors encoded implicitly. But it still remains unclear if these policy functions can encode semantic priors when learned via RL.

\begin{figure*}
\centering
\begin{minipage}{0.99\textwidth}
\centering
\includegraphics[width=\linewidth,height=\textheight,keepaspectratio]{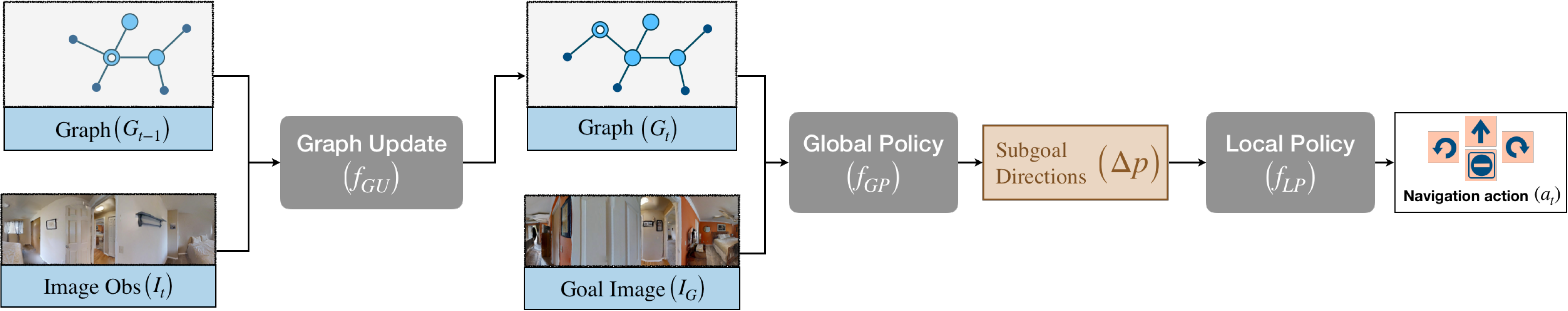}
\end{minipage}
\vspace{3pt}

    \caption{\small \textbf{Model Overview.} Figure showing an overview of the proposed model, \methodname. It consists of 3 components, a Graph Construction module which updates the topological map as it receives observations, a Global Policy which samples subgoals, and a Local Policy which takes navigational actions to reach the subgoal. See text for more details.}
\label{fig:overview}
\end{figure*}

In this paper, we propose to tackle both the problems head-on. Instead of using metric-maps which are brittle to localization and noise, we propose a topological representation of the space. Our proposed representation consists of nodes that are connected in the form of a graph, based on local geometry information. Each node is represented visually via a 360-degree panoramic image. Nodes are connected to each other using approximate relative pose between them. But what makes our visual topological maps novel are two directional functions~$\mathcal{F}_g$ and $\mathcal{F}_s$, which extract geometric and semantic properties of nodes. Specifically, $\mathcal{F}_g$ estimates how likely agent will encounter free-space and $\mathcal{F}_s$ estimates how likely target image is to be encountered if the agent moves in a particular direction. By explicitly modeling and learning function $\mathcal{F}_s$, our model ensures that structural priors are encoded and used when exploring and navigating new unseen environments. 

Our representation has few advantages over classical and end-to-end learning-based approaches:  (a) it uses graph-based representation which allows efficient long-term planning; (b) it explicitly encodes structural priors via function $\mathcal{F}_s$; (c) the geometric function $\mathcal{F}_g$ allows efficient exploration and online map building for a new environment; (d) but most importantly, all the functions and policies can be learned in completely supervised manner forgoing the need for unreliable credit assignment via RL.

\section{Related Work}
\seclabel{related}
\noindent Our paper makes contributions across the following multiple aspects of the navigation problem: space representation, training paradigm for navigation policies, and different navigation tasks. We survey works in these areas below.\\\vspace{-6pt}

\noindent \textbf{Navigation Tasks.} 
Navigation tasks can be divided into two main categories. The first category of tasks is ones where the goal location is known, and limited exploration is necessary. This could be in the form of a simply wandering around without colliding~\cite{gandhi2017learning,sadeghi2016cad2rl}, following an object~\cite{lee2017learning}, getting to a goal coordinate~\cite{gupta2017cognitive,anderson2018evaluation}: using sequence of images along the path~\cite{kumar2018visual,bruce2018deployable}, or language-based instructions~\cite{anderson2018vision}. Sometimes, the goal is specified as an image but experience from the environment is available in the form of demonstrations~\cite{eysenbach2019search,savinov2018semi}, or in the form of reward-based training~\cite{mirowski2018learning,zhu2017target}, which again limits the role of exploration. The second category of tasks is when the goal is not known and exploration is necessary. Examples are tasks such as finding an object~\cite{gupta2017cognitive}, or room~\cite{wu2019bayesian}, in a novel environment, or explicit exploration~\cite{chen2018learning,anm}. These task categories involve different challenges. The former tasks focus on effective retrieval and robust execution, while the later tasks involve semantic and common sense reasoning in order to efficiently operate in previously unseen environments. Our focus, in this work, is the task of reaching a target image in a novel environment. No experience is available from the environment, except for the target image. We aren't aware of any works that target this specific problem.\\\vspace{-6pt}

\noindent \textbf{Classical Space Representations.}
Spatial and topological representations have a rich history in robot navigation. Researchers have used explicit metric spatial representations~\cite{elfes1989using}, and have considered how can such representations be built with different sensors~\cite{newcombe2011kinectfusion, thrun2005probabilistic, newcombe2011dtam, mur2015orb, hornung13auro}, and how can agents be localized against such representations~\cite{dellaert1999monte}. Recent work has started associating semantics with such spatial representations~\cite{bowman2017probabilistic}. In a similar vein, non-metric topological representations have also been considered in classical literature~\cite{kuipers1991robot, meng1993mobile, choset2001topological}. Some works combine topological and metric representations~\cite{tomatis2001combining, thrun1998integrating}, and some study topological representations that are semantic~\cite{kuipers1991robot}. While our work builds upon the existing literature on topological maps, the resemblance is only at high-level graph structure. Our work focuses on making visual topological mapping and exploration scalable, robust and efficient. We achieve this via the representation of both semantic and geometric properties in our topological maps; the ability to build topological maps in an online manner and finally, posing the learning problem as a supervised problem.\\\vspace{-6pt}

\noindent \textbf{Learned Space Representations.}
Depending on the problem being considered, different representations have been investigated. For short-range locomotion tasks, purely reactive policies~\cite{sadeghi2016cad2rl, gandhi2017learning, lee2017learning, bansal2019combining} suffice. For more complex problems such as target-driven navigation in a novel environment, such purely reactive strategies do not work well~\cite{zhu2017target}, and memory-based policies have been investigated. This can be in the form of vanilla neural network memories such as LSTMs~\cite{mirowski2016learning, oh2016control}, or transformers~\cite{fang2019smt}. Researchers have also incorporated insights from classical literature into the design of expressive neural memories for navigation. This includes spatial memories~\cite{gupta2017cognitive, parisotto2017neural} and topological approaches~\cite{chen2019behavioral, savinov2018semi, eysenbach2019search, wu2019bayesian, savinov2018episodic, yang2018visual}. 
Learned spatial approaches can acquire expressive spatial representations~\cite{gupta2017cognitive}, they are however bottle-necked by their reliance on metric consistency and thus have mostly been shown to work in discrete state spaces for comparatively short-horizon tasks~\cite{gupta2017cognitive, parisotto2017neural}. Researchers have also tackled the problem of passive and active localization~\cite{mur2015orb,chaplot2018active}, in order to aid building such consistent metric representations. 
Some topological approaches \cite{chen2019behavioral, savinov2018semi, eysenbach2019search} work with human explorations or pre-built topological maps, thus ignoring the problem of exploration. Others build a topological representation with explicit semantics~\cite{wu2019bayesian, zhang2018composable}, which limits tasks and environments that can be tackled. In contrast from past work, we unify spatial and topological representations in a way that is robust to actuation error, show how we can incrementally and autonomously build topological representations, and do semantic reasoning.\\
\vspace{-6pt}

\begin{figure}[t]
\centering
\begin{minipage}{0.99\linewidth}
\centering
\includegraphics[width=\linewidth,height=\textheight,keepaspectratio]{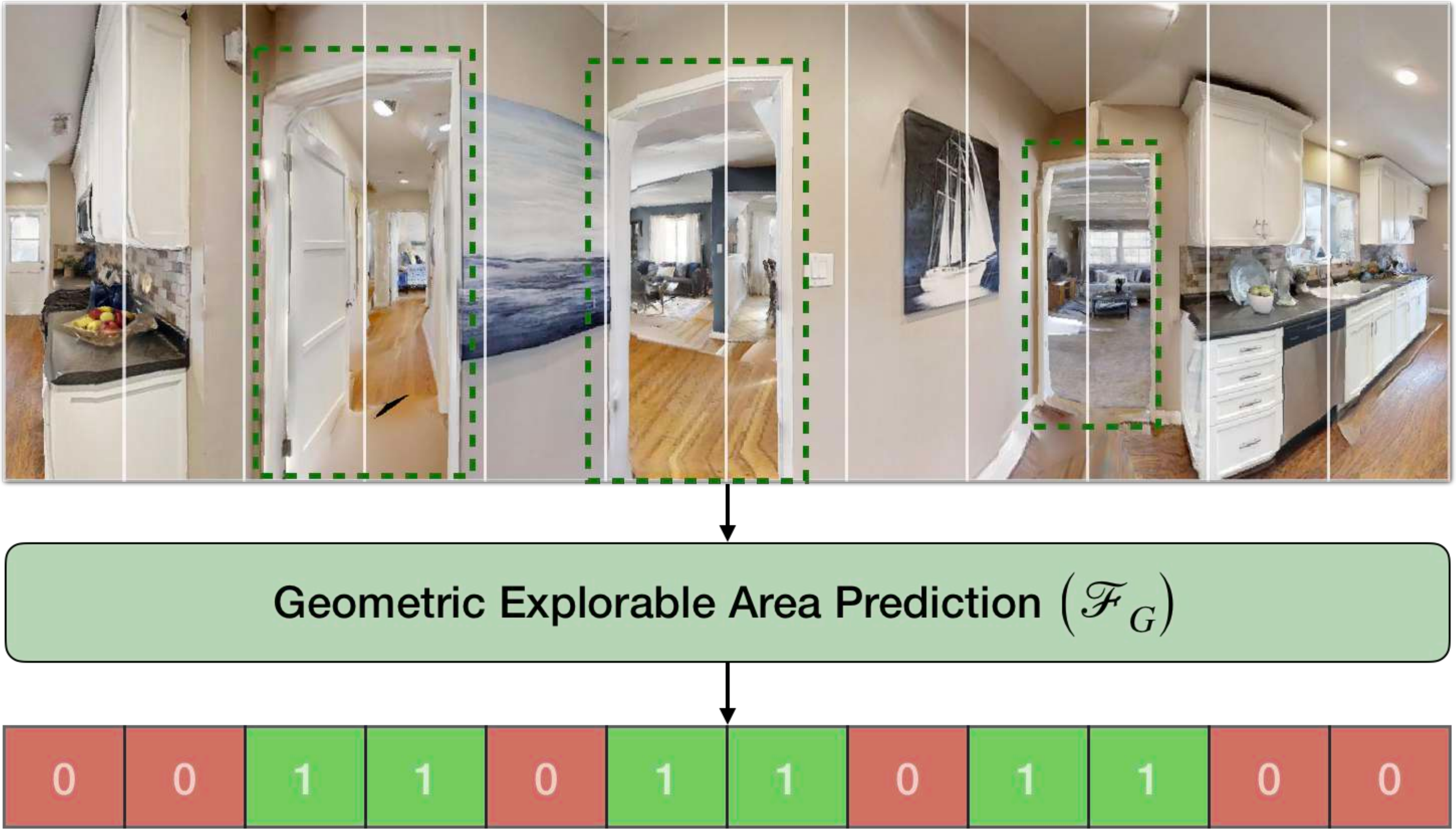}
\end{minipage}

    \caption{\small \textbf{Geometric Explorable Area Prediction ($\mathcal{F}_G$).} Figure showing sample input image ($I$) and output predictions of the Geometric Explorable Area Prediction function ($\mathcal{F}_G$). The green boxes show doorways for reference, and are not available as input.
    }
\label{fig:f_g}
\end{figure}

\noindent \textbf{Training Methodology.}
Different tasks have also lead to the design of different training methodologies for training navigation policies. This ranges from reinforcement learning with sparse and shaped rewards~\cite{mirowski2018learning, zhu2017target, pathakICMl17curiosity, sadeghi2016cad2rl, mirowski2016learning}, imitation learning and DAgger~\cite{gupta2017cognitive, ross2011reduction}, self-supervised learning for individual components~\cite{savinov2018semi,gandhi2017learning}. While RL allows learning of rich exploratory behavior, training policies using RL is notoriously hard and sample inefficient. Imitation learning is sample efficient but may not allow learning exploratory behavior. Self-supervised learning is promising but has only been experimented in the context of known goal tasks. We employ a supervised learning approach and show how we can still learn expressive exploration behavior while at the same time not suffering from exuberant sample complexity for training.
\vspace{10pt}

\begin{figure}[t]
    \centering
\begin{minipage}{0.99\linewidth}
\centering
    \includegraphics[width=\linewidth,height=\textheight,keepaspectratio]{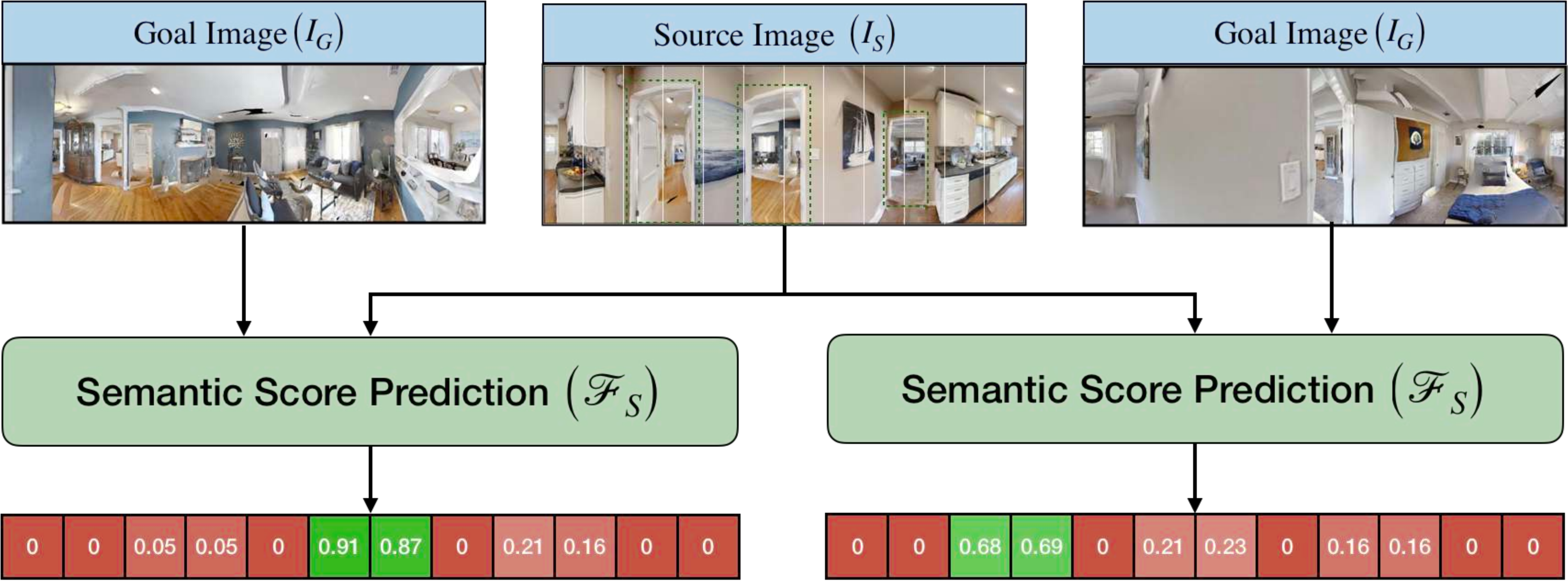}
\end{minipage}
\vspace{3pt}

    \caption{\small \textbf{Semantic Score Prediction ($\mathcal{F}_S$).} Figure showing sample input and output predictions of the Semantic Score Prediction function ($\mathcal{F}_S$). The score predictions change based on the goal image. When the goal image is of the living room (left), the score of the directions in the center are higher as they lead to the living room. When the goal image is of a bedroom (right), the scores corresponding to the pathway on the left are higher as they are more likely to lead to the bedroom.}
    \label{fig:f_s}
\end{figure}

\section{Task Setup}
\noindent We consider an autonomous agent situated in an episodic environment. At the beginning of an episode, the agent receives a target goal image, $I_G$. At each time step $t$, the agent receives observations ($s_t$) from the environment. Each observation consists of the current first-person image observation, $I_t$, from a panoramic camera and a pose estimate from a noisy motion sensor. At each time step, the agent takes a navigational action $a_t$. The objective is to learn a policy $\pi(a_t| s_t, I_G)$ to reach the goal image. In our experimental setup, all images are panoramas, including agent observations and goal image.

\section{Methods}
\noindent We propose a modular model, `\methodname~(\methodabbr)', which builds and maintains a topological map for navigation. The topological map is represented using a graph, denoted by $G_t$ at time $t$. Each node in the graph ($N_i$) is associated with a panoramic image ($I_{N_i}$) and represents the area visible in this image. Two nodes are connected by an edge ($E_{i,j}$) if they represent adjacent areas. Each edge also stores the relative pose between two nodes, $\Delta p_{ij}$.

Our model consists of three components, a \textit{Graph  Update} module, a \textit{Global Policy}, and a \textit{Local Policy}. On a high level, the Graph Update module updates the topological map based on agent observations, the Global Policy selects a node in the graph as the long-term goal and finds a subgoal to reach the goal using path planning, and the Local Policy navigates to the subgoal based on visual observations. Fig.~\ref{fig:overview} provides an overview of the proposed model.

\begin{figure*}
\centering
\begin{minipage}{0.99\textwidth}
\centering
\includegraphics[width=\linewidth,height=\textheight,keepaspectratio]{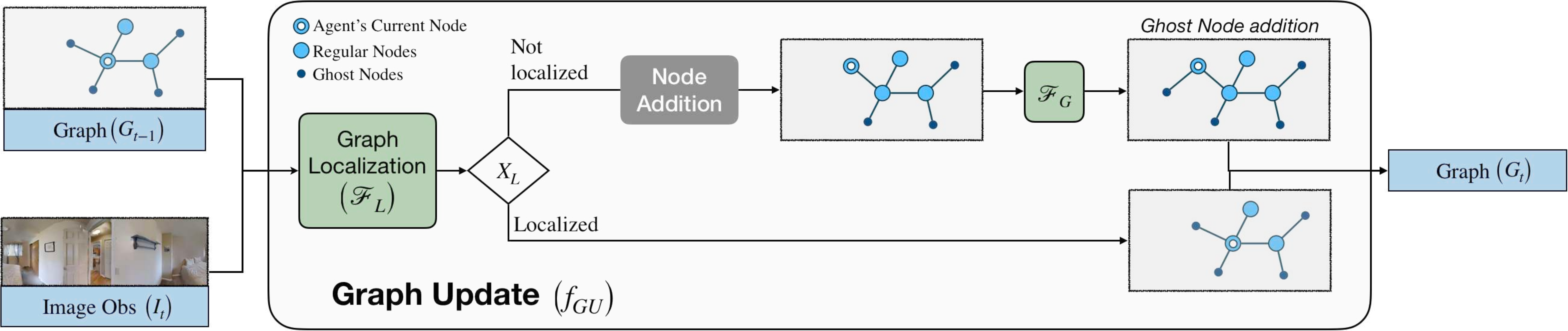}
\end{minipage}
\vspace{3pt}

    \caption{\small \textbf{Graph Update.} Figure showing an overview of the Graph Update Module. It takes the current Graph ($G_t$) and the agent observation ($I_t$) as input. It first tries to localize the agent in the Graph. If the agent is localized in a node different from the last timestep, it changes the location of the agent and adds an edge if required. If the agent is not localized, a new node is added and corresponding ghost nodes are added using the geometric explorable area prediction function ($\mathcal{F}_G$). See the text for more details.}
\label{fig:update}
\end{figure*}

The above components will require access to 4 functions. We first define these 4 functions and then describe how they are used by the components of the model.\\\vspace{-4pt}

\noindent \textbf{Graph Localization $(\mathcal{F}_L)$.} Given a graph $G$ and an image $I$, this function tries to localize $I$ in a node in the graph. An image is localized in a node if its location is visible from the panoramic image associated with the node. Internally, this requires comparing each node image ($I_{N_i}$) with the given image ($I$) to predict whether $I$ belongs to node $N_i$.\\\vspace{-4pt}

\noindent \textbf{Geometric Explorable Area Prediction $(\mathcal{F}_G)$}. Given an image $I$, this function makes $n_\theta = 12$ different predictions of whether there's explorable area in the direction $\theta$ sampled uniformly between $0$ and $2\pi$. Figure~\ref{fig:f_g} shows an example of input and output of this function. Intuitively, it recognizes doors or hallways which lead to other areas.\\\vspace{-4pt}

\noindent \textbf{Semantic Score Prediction $(\mathcal{F}_S)$}. Given a source image $I_S$ and a goal image $I_G$, this function makes $n_\theta = 12$ different predictions of how fast the agent is likely to reach the goal image if it explores in the direction $\theta$ sampled uniformly between $0$ and $2\pi$. Figure~\ref{fig:f_s} shows example input-output pairs for this function. The scores corresponding to the same source image change as the goal image changes. Estimating this score requires the model to learn semantic priors about the environment. \\\vspace{-4pt}

\noindent \textbf{Relative Pose Prediction $(\mathcal{F}_R)$}. Given a source image $I_S$ and a goal image $I_G$ which belong to the same node, this function predicts the relative pose ($\Delta p_{S,G}$) of the goal image from the source image.\\\vspace{-6pt}

\subsection{Model components}
\vspace{-2pt}
\noindent Assuming that we have access to the above functions, we first describe how these functions are used by the three components to perform Image Goal Navigation. We then describe how we train a single model to learn all the above functions using supervised learning.\\\vspace{-4pt}

\noindent \textbf{Graph Update.} The Graph Update module is responsible for updating the topological map given agent observations. At $t=0$, the agent starts with an empty graph. At each time step, the Graph Update module ($f_{GU}$) takes in the current observations $s_t$ and the previous topological map $G_{t-1}$ and outputs the updated topological map, $G_{t} = f_{GU}(s_t, G_{t-1})$. Figure~\ref{fig:update} shows an overview of the Graph Update Module.

In order to update the graph, the module first tries to localize the current image in a node in the current graph using the Graph Localization function ($\mathcal{F}_L$). If the current image is localized in a node different from the last node, we add an edge between the current node and the last node (if it does not exist already). If the current image is not localized, then we create a new node with the current image. We also add an edge between the new node and the last node. Every time we add an edge, we also store the relative pose ($\Delta p$) between the two nodes connected by the edge using the sensor pose estimate. 

The above creates a graph of the explored areas. In order to explore new areas, we also predict and add unexplored areas to the graph. We achieve this by augmenting the graph with `ghost' nodes which are agent's prediction of explorable areas using the Geometric Explorable Area Prediction function ($\mathcal{F}_G$). If there is an explorable area in a direction $\theta$, we add a `ghost' node ($X_k$) and connect it to the new node ($N_i$) using edge $E_{i,k}$. Since we do not have the image at the ghost node location, we associate a patch of the node image in the direction of $\theta$, i.e. ($I_{X_k} = I_{{N_i},\theta}$). The relative pose between the new node and the ghost node is stored as ($r, \theta$), where $\theta$ is the direction and $r = 3m$ is the radius of the node. The ghost nodes are always connected to exactly one regular node and always correspond to unexplored areas. We ensure this by removing ghost nodes when adding regular nodes in the same direction, and not adding ghost nodes in a particular direction if a regular node exists in that direction. Intuitively, ghost nodes correspond to the unexplored areas at the boundary of explored areas denoted by regular nodes. \\\vspace{-4pt}

\begin{figure*}
\centering
\begin{minipage}{0.99\textwidth}
\centering
\includegraphics[width=\linewidth,height=\textheight,keepaspectratio]{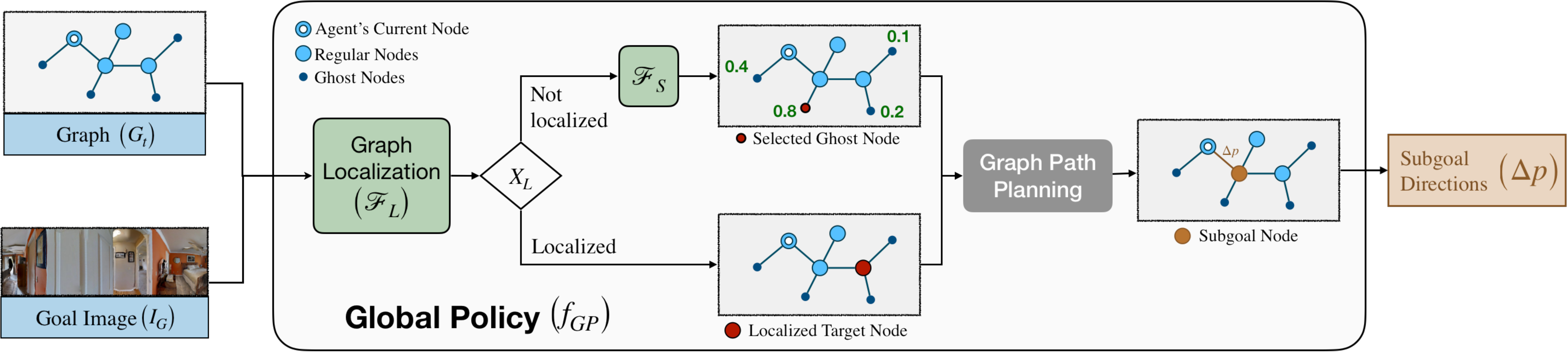}
\end{minipage}
\vspace{3pt}

    \caption{\small \textbf{Global Policy.} Figure showing an overview of the Global Policy. It takes the current Graph ($G_t$) and the Goal Image ($I_G$) as input. It first tries to localize the Goal Image in the Graph. If the Goal Image is localized, the corresponding node is selected as the long-term goal. If the Goal Image is not localized, then the semantic scoring function ($\mathcal{F}_S$) is used to score all the ghost nodes based on how close they are to the goal image. The ghost node with the highest score is selected as the long-term goal. Given a long-term goal, a subgoal node is computed using graph path planning. The relative directions to the subgoal node are the output of the Global Policy which is passed to the Local Policy.}
\label{fig:global}
    \vspace{-8pt}
\end{figure*}

\noindent\textbf{Global Policy.}
The Global Policy is responsible for selecting a node in the above graph as the long-term goal. It first tries to localize the goal image in the current graph using the Graph Localization function ($\mathcal{F}_L$). If the goal image ($I_G$) is localized in a node $N_i$, then $N_i$ is chosen as the long-term goal. If the goal image is not localized, then the Global Policy needs to choose an area to explore, i.e. choose a ghost node for exploration. We use the Semantic Score Prediction ($\mathcal{F}_S$) function to predict the score of all ghost nodes. The Global Policy then just picks the ghost node with the highest score as the long-term goal.

 Once a node is selected as a long-term goal, we plan the path to the selected node from the current node using Djikstra's algorithm on the current graph. The next node on the shortest path is chosen to be the subgoal ($N_{SG}$). The relative pose associated with the edge to the subgoal node is passed to the Local Policy ($\Delta p_{i, SG}$).

If the goal image is localized in the current node (or the agent reaches the node $N_i$ where the goal image ($I_G$) is localized), the Global policy needs to predict the relative pose of $I_G$ with respect to the current agent observation. We use the Relative Pose Prediction ($\mathcal{F}_R$) function to get the relative pose of the goal image, which is then passed to the Local Policy.\\\vspace{-4pt}

\noindent\textbf{Local Policy.}
The Local Policy receives the relative pose as goal directions which comprises of distance and angle to the goal. Given the current image observation and the relative goal directions, the Local Policy takes navigation actions to reach the relative goal. This means the Local Policy is essentially a PointGoal navigation policy. Our Local Policy is adapted from~\cite{anm}. It predicts a local spatial map using a learned mapper model in the case of RGB input or using geometric projections of the depth channel in case of RGBD input. It then plans a path to the relative goal using shortest path planning.

\begin{figure*}
\centering
\begin{minipage}{0.99\textwidth}
\centering
\includegraphics[width=\linewidth,height=\textheight,keepaspectratio]{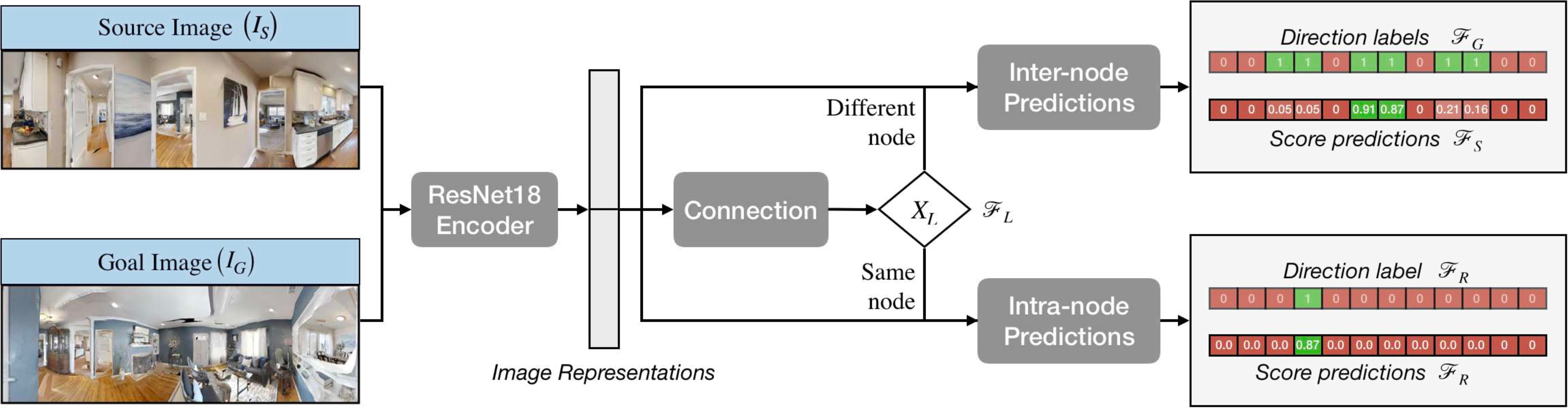}
\end{minipage}
\vspace{3pt}

    \caption{\small \textbf{NTS Multi-task Learning Model.} Figure showing an overview of the NTS Multi-task Learning Model. It takes a Source Image ($I_S$) and a Goal Image ($I_G$) as input and encodes them using a shared ResNet18 encoder. It first predicts whether the two images belong to the same node or not. If they belong to the same node, it makes Intra-Node predictions which include the direction and score (or equivalently) distance of the Goal Image relative to the Source Image. If they belong to different nodes, it makes Inter-Node Predictions which include directions of explorable areas and a semantic score corresponding to each explorable area denoting its proximity of the Goal Image. All the predictions of this model are used at various places in the components of the overall NTS model. See text for more details.}
\label{fig:model}
\end{figure*}

\vspace{2pt}
\subsection{Training NTS Multi-task Learning model}
\vspace{-2pt}
Given access to the four functions described above, we discussed how the different components use these functions for navigation. In this subsection, we describe how we train a single multi-task learning model to learn all the four functions. Figure~\ref{fig:model} shows an overview of this multi-task learning model. It takes a Source Image ($I_S$) and a Goal Image ($I_G$) as input and encodes them using a shared ResNet18~\cite{he2016deep} encoder. It first predicts whether the two images belong to the same node or not. This prediction is used to implement the Graph Localization function ($\mathcal{F}_L$). If they belong to the same node, it makes Intra-Node predictions which include the direction and score (or equivalently) distance of the Goal Image relative to the Source Image. These predictions are used to implement the Relative Pose Prediction function ($\mathcal{F}_R$). If they belong to different nodes, it makes Inter-Node Predictions which include directions of explorable areas (which is used as the Geometric Explorable Area Prediction function ($\mathcal{F}_G$)) and a semantic score corresponding to each explorable area denoting its proximity of the Goal Image (which is used as the Semantic Score Prediction function ($\mathcal{F}_S$)). The Connection, Intra-Node Prediction, and Inter-Node Prediction models consist of fully-connected layers with ReLU activations and dropout. Exact details are deferred to the supplementary material.

\section{Experimental Setup}
\vspace{-4pt}
\noindent\textbf{Environment.} All our experiments are conducted in the Habitat simulator~\cite{savva2019habitat} with the Gibson~\cite{xiazamirhe2018gibsonenv} dataset. The Gibson dataset is visually realistic as it consists of reconstructions of real-world scenes. We also implement physically realistic motion sensor and actuation noise models as proposed by~\cite{anm}. Actuation motion noise leads to stochastic transitions as the amount translated or rotated by the agent is noisy. This model also adds realistic translational noise in rotation actions and rotational noise in translational actions. The sensor noise model adds realistic noise to the base odometry sensor readings. Both the noise models are based on real-world data and agents trained on these noise models are shown to transfer to the real-world~\cite{anm}. 

\noindent\textbf{Task setup.} We use panoramic images of size $128 \times 512$ for both agent image observation and target goal image. We conduct experiments with both RGB and RGBD settings. The base odometry sensor provides a $3 \times 1$ reading denoting the change in agent's x-y coordinates and orientation. The action space consists of four actions: \texttt{move\_forward, turn\_right, turn\_left, stop}. The forward action moves the agent approximately 25cm forward and the turn actions turn the agent approximately 10 degrees. Note that the state space and motion of the agent are continuous. The agent succeeds in an episode if it takes the \texttt{stop} action within a $1m$ radius of the target location. The agent fails if it takes \texttt{stop} action anywhere else or does not take the \texttt{stop} till the episode ends. In addition to the success rate, we also use Success weighted by inverse Path Length (SPL) as an evaluation metric as proposed by \cite{anderson2018evaluation}. It takes into account the efficiency of the agent in reaching the goal (shorter successful trajectories lead to higher SPL). \\\vspace{-6pt}

\noindent\textbf{Training data.} We split the curated set of 86 scenes from ~\cite{savva2019habitat} into sets of 68/4/14 scenes for train/val/test. For training our supervised learning model, we sample 300 images randomly in each of 68 training scenes. We get labels for pairs of source and target images in each scene giving us a total of approximately $ 68 \times 300 \times 300$ = 6.12 million data points. The labeling process is automated and it only requires the ground-truth map already available with the dataset without the need for any additional human annotation. Details of the labeling process are deferred to the supplementary material. Note that sampling images or the ground-truth map are not required for the test environments.\\\vspace{-6pt}

\noindent\textbf{Test episodes.} For creating test episodes, we sample episodes (given by starting and goal locations) in the test scenes to create 3 different sets of difficulty based on the distance of the goal from starting locations: Easy ($1.5-3m$), Medium ($3-5m$) and Hard ($5-10m$). The maximum episode length is 500 steps for each difficulty level. 

\setlength{\tabcolsep}{6.8pt}
\begin{table*}[]
    \centering
\begin{tabular}{@{}llccccccccccc@{}}
\toprule
\multicolumn{1}{c}{}  &                                           & \multicolumn{2}{c}{Easy}      &           & \multicolumn{2}{c}{Medium}    &           & \multicolumn{2}{c}{Hard}      &           & \multicolumn{2}{c}{Overall}   \\
\multicolumn{1}{c}{}  & Model                                     & Succ          & SPL           &           & Succ          & SPL           &           & Succ          & SPL           &           & Succ          & SPL           \\ \midrule
\multirow{4}{*}{RGB}  & ResNet + GRU + IL                        & 0.57          & 0.23          &           & 0.14          & 0.06          &           & 0.04          & 0.02          &           & 0.25          & 0.10          \\
                      & Target-driven RL~\cite{zhu2017target}                  & 0.56          & 0.22          &           & 0.17          & 0.06          &           & 0.06          & 0.02          &           & 0.26          & 0.10          \\
                      & Active Neural SLAM (ANS)~\cite{anm}       & 0.63          & 0.45          &           & 0.31          & 0.18          &           & 0.12          & 0.07          &           & 0.35          & 0.23          \\
                      & \textbf{Neural Topological SLAM (NTS)} & \textbf{0.80} & \textbf{0.60} & \textbf{} & \textbf{0.47} & \textbf{0.31} & \textbf{} & \textbf{0.37} & \textbf{0.22} & \textbf{} & \textbf{0.55} & \textbf{0.38} \\ \midrule
\multirow{6}{*}{RGBD} & ResNet + GRU + IL                           & 0.72          & 0.32          &           & 0.16          & 0.09          &           & 0.05          & 0.02          &           & 0.31          & 0.14          \\
                      & Target-driven RL~\cite{zhu2017target}        & 0.68          & 0.28          &           & 0.21          & 0.08          &           & 0.09          & 0.03          &           & 0.33          & 0.13          \\
                      & Metric Spatial Map + RL~\cite{chen2018learning}     & 0.69          & 0.27          &           & 0.22          & 0.07          &           & 0.12          & 0.04          &           & 0.34          & 0.13          \\
                      & Metric Spatial Map + FBE + Local                       & 0.77          & 0.56          &           & 0.36          & 0.18          &           & 0.13          & 0.05          &           & 0.42          & 0.26          \\
                      & Active Neural SLAM (ANS)~\cite{anm}       & 0.76          & 0.55          &           & 0.40          & 0.24          &           & 0.16          & 0.09          &           & 0.44          & 0.29          \\
                      & \textbf{Neural Topological SLAM (NTS)} & \textbf{0.87} & \textbf{0.65} & \textbf{} & \textbf{0.58} & \textbf{0.38} & \textbf{} & \textbf{0.43} & \textbf{0.26} & \textbf{} & \textbf{0.63} & \textbf{0.43} \\ \bottomrule
\end{tabular}
\vspace{-2pt}
\caption{\textbf{Results.} Performance of the proposed model Neural Topological SLAM (NTS) and the baselines in RGB and RGBD settings.}
\vspace{-8pt}
\label{tab:results_main}
\end{table*}

\subsection{Baselines}
\noindent We use the following baselines for our experiments:

\noindent\textbf{ResNet + GRU + IL.} A simple baseline consisting of ResNet18 image encoder and GRU based policy trained with imitation learning (IL).

\noindent\textbf{Target-driven RL.} A siamese style model for encoding the current image and the goal image using shared convolutional networks and trained end-to-end with reinforcement learning, adapted from Zhu et al.~\cite{zhu2017target}.

\noindent\textbf{Metric Spatial Map + RL.} An end-to-end RL model which uses geometric projections of the depth image to create a local map and passes it to the RL policy, adapted from Chen et al.~\cite{chen2018learning}.

\noindent\textbf{Metric Spatial Map + FBE + Local.} This is a hand-designed baseline which creates a map using depth images and then uses a classical exploration heuristic called Frontier-based Exploration (FBE)~\cite{yamauchi1997frontier} which greedily explores nearest unexplored frontiers in the map. We use the Localization model and Local Policy from NTS to detect when a goal is nearby and navigate to it.

\noindent\textbf{Active Neural SLAM.} This is a recent modular model based on Metric Spatial Maps proposed for the task of exploration. We adapt it to the Image Goal task by using the Localization model and Local Policy from NTS to detect when a goal is nearby and navigate to it.

All the baselines are trained for 25 million frames. RL baselines are trained using Proximal Policy Optimization~\cite{schulman2017proximal} with a dense reward function. The reward function includes a high reward for success (=SPL*10.), a shaping reward equal to the decrease in distance to the goal and a per step reward of -0.001 to encourage shorter trajectories. ResNet + GRU + IL is trained using behavioral cloning on the ground-truth trajectory. This means just like the proposed model, all the baselines also use the ground-truth map for training. In terms of the number of training samples, sampling 300 random images in the environment would require 300 episode resets in the RL training setup. 300 episodes in 68 scenes would lead to a maximum of $10.2$ million ($=68 \times 300 \times 500$) samples. Since we use 25 million frames to train our baselines, they use strictly more data than our model. Furthermore, our model does not require any interaction in the environment and can be trained offline with image data.

\setlength{\tabcolsep}{8.2pt}
\begin{table*}[]
    \centering
\begin{tabular}{@{}lcccccccccc@{}}
\toprule
                                        &  & \multicolumn{4}{c}{RGBD - No stop} &  & \multicolumn{4}{c}{RGBD - No Noise} \\
Model                              &  & Easy   & Med.   & Hard  & Overall  &  & Easy   & Med.   & Hard  & Overall  \\ \midrule
ResNet + GRU + IL                     &  & 0.76   & 0.28   & 0.10  & 0.38     &  & 0.71   & 0.18   & 0.06  & 0.32     \\
Target-driven RL~\cite{zhu2017target}               &  & 0.89   & 0.45   & 0.21  & 0.52     &  & 0.69   & 0.22   & 0.07  & 0.33     \\
Metric Spatial Map + RL~\cite{chen2018learning}               &  & 0.89   & 0.45   & 0.21  & 0.52     &  & 0.70   & 0.24   & 0.11  & 0.35     \\
Metric Spatial Map + FBE + Local                 &  & 0.92   & 0.46   & 0.29  & 0.56     &  & 0.78   & 0.46   & 0.23  & 0.49     \\
Active Neural SLAM (ANS)~\cite{anm}  &  & 0.93   & 0.50   & 0.32  & 0.58     &  & 0.79   & 0.53   & 0.30  & 0.54     \\
    \textbf{Neural Topological SLAM (NTS)}    &  & \textbf{0.94}   & \textbf{0.70}   & \textbf{0.60}  & \textbf{0.75}     &  & \textbf{0.87}   & \textbf{0.60}   & \textbf{0.46}  & \textbf{0.64}     \\ \bottomrule
\end{tabular}
\vspace{-2pt}
    \caption{\textbf{No stop and no noise.} Success rate of all the models without stop action (left) and without motion noise (right).}
\vspace{-10pt}
\label{tab:results_stop}
\end{table*}

\section{Results}
\begin{figure*}[h!]
\centering
\includegraphics[width=\linewidth,height=\textheight,keepaspectratio]{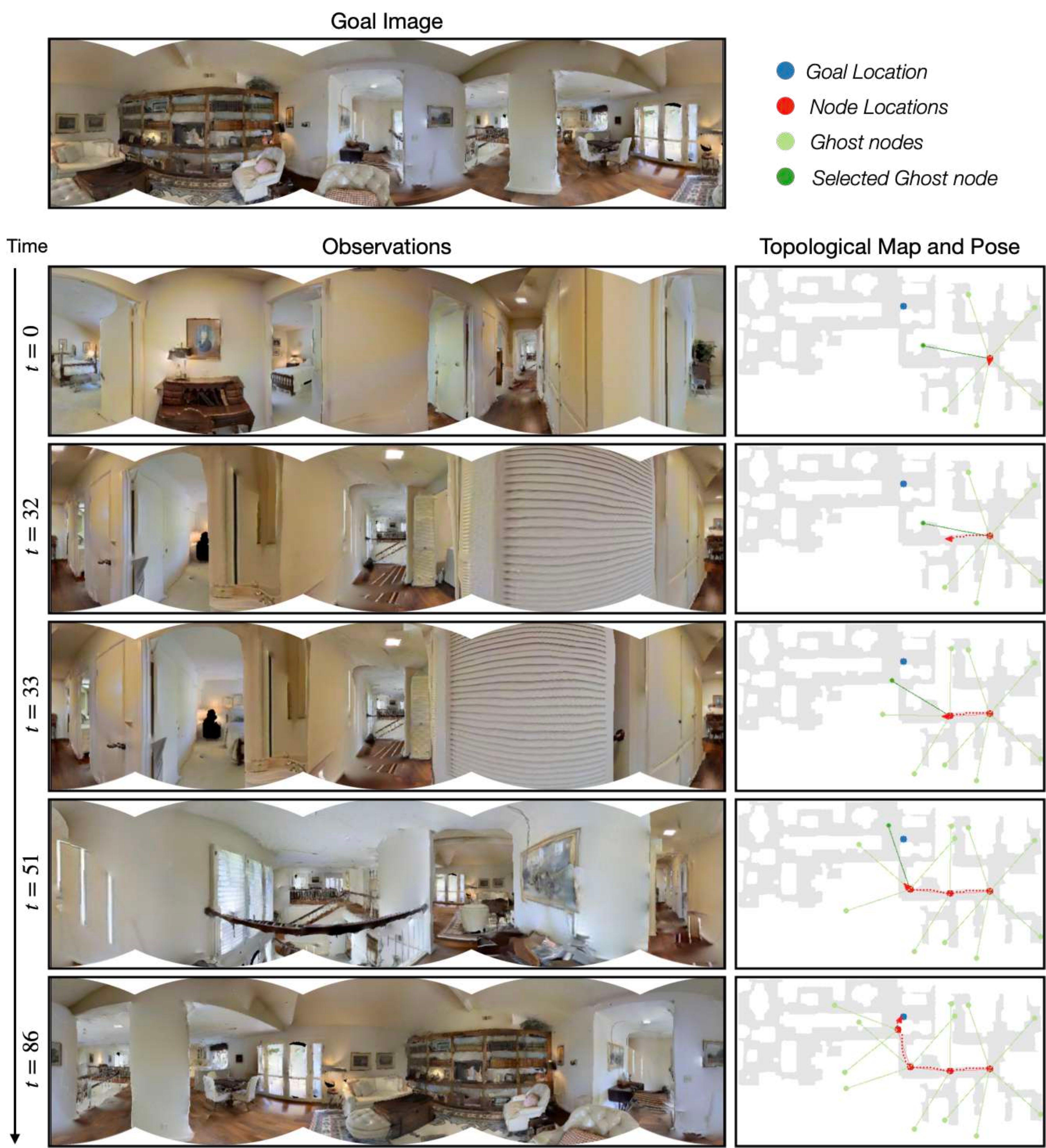}
	\vspace{-6pt}
	\caption{\small \textbf{Example Trajectory Visualization.} Figure showing a visualization of an example trajectory using the NTS model. Agent observations are shown on the left and the topological map and pose are shown on the right. Note that the map and the goal location are shown only for visualization and are not available to the NTS model. \textbf{(t=0)} NTS model creates the first node and adds the ghost nodes. Note that ghost node locations are not predicted, the model predicts only the direction of ghost nodes. The model correctly selects the correct ghost node closest to the goal. \textbf{(t=33)} The NTS model creates the second regular node and adds ghost nodes. Note that the ghost node in the direction of the second node from the first node are removed. \textbf{(t=51)} The NTS model creates another new node. \textbf{(t=86)} The agent reaches the goal after creating 4 nodes as shown in the trajectory on the map and decides to take the stop action.
    }
\label{fig:eg}
\end{figure*}

\vspace{-4pt}
\noindent We evaluate the proposed method and all the baselines on 1000 episodes for each difficulty setting. We compare all the methods across all difficulty levels in both RGB and RGBD settings in Table~\ref{tab:results_main}. The results show that the proposed method outperforms all the baselines by a considerable margin across all difficulty settings with an overall Succ/SPL of 0.55/0.38 vs 0.35/0.23 in RGB and 0.63/0.43 vs 0.44/0.29 in RGBD. The results also indicate the relative improvement of \methodabbr{} over the baselines increases as the difficulty increases leading to a large improvement in the hard setting (0.43/0.26 vs 0.16/0.09 in RGBD). In Fig~\ref{fig:eg}, we visualize an example trajectory using the NTS model.
\\\vspace{-6pt}

\begin{figure*}
\centering
\begin{minipage}{0.49\textwidth}
\includegraphics[width=\linewidth,height=\textheight,keepaspectratio]{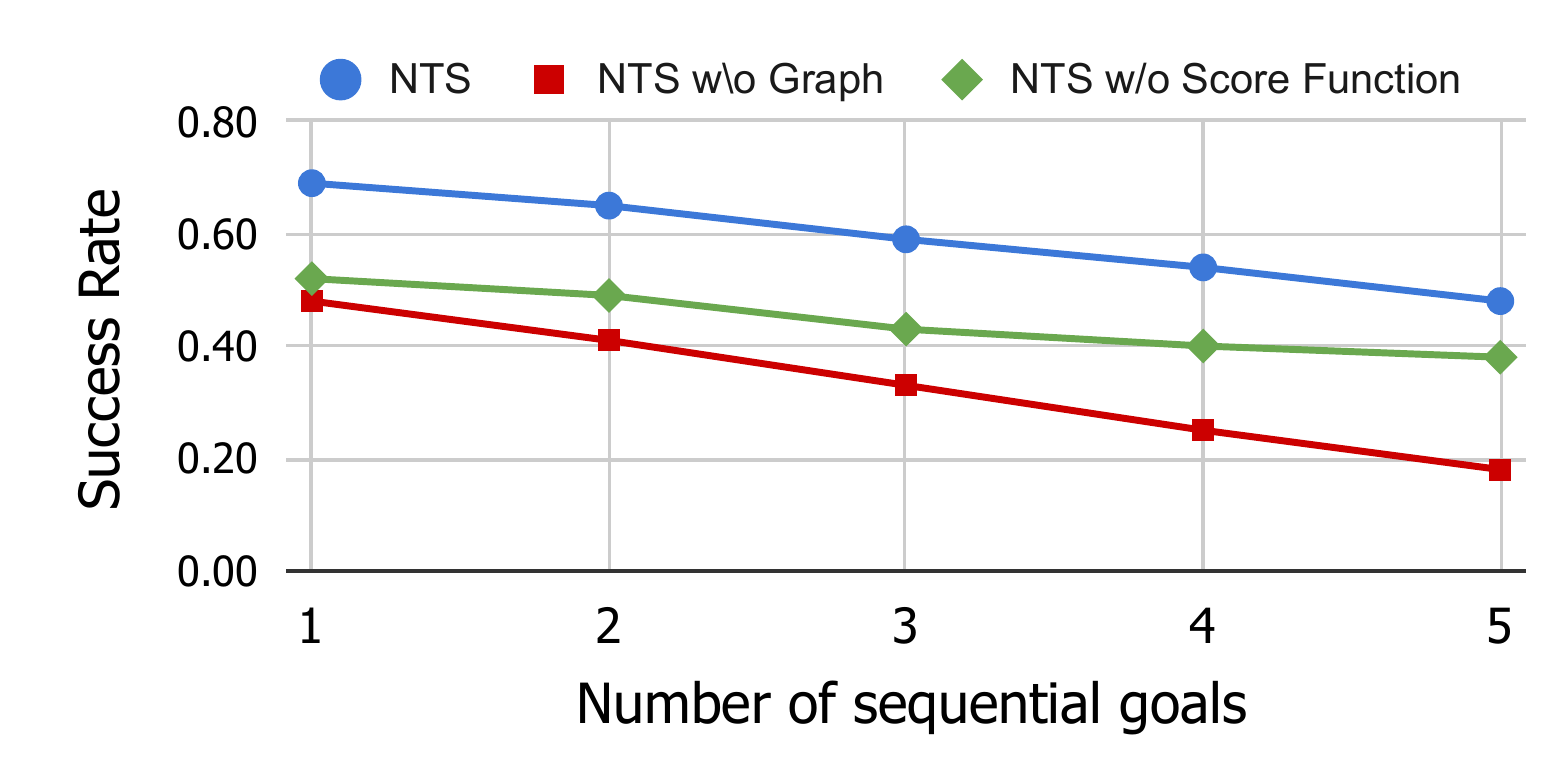}
\end{minipage}\hfill
\begin{minipage}{0.49\textwidth}
\centering
\includegraphics[width=\linewidth,height=\textheight,keepaspectratio]{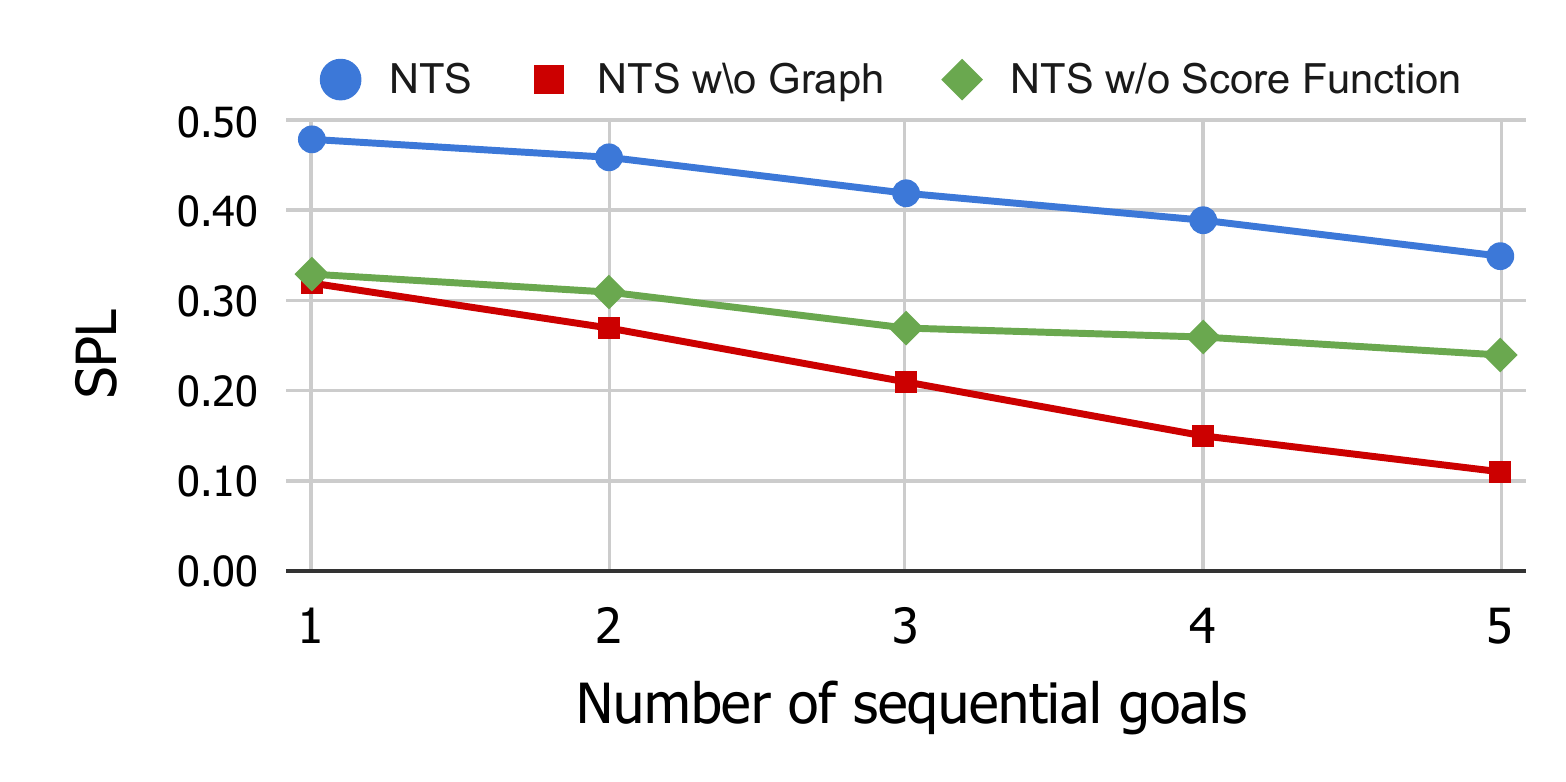}
\end{minipage}
\vspace{-8pt}
\caption{\small Performance of the proposed model \methodabbr{} and two ablations as a function of number of sequential goals. }
\label{fig:ablation}
\vspace{-12pt}
\end{figure*}

\noindent \textbf{Comparison with end-to-end RL and the effect of stop action.} The results indicate that \methodabbr{} performs better than both end-to-end RL based baselines~\cite{zhu2017target, chen2018learning} and methods using metric spatial maps~\cite{chen2018learning, anm}. The performance of the RL-based baselines is much weaker than the proposed model. We believe the reason behind this is the complexity of the exploration search space. Compared to the Pointgoal navigation task where the agent receives updated direction to the goal at each time step, Image Goal navigation is more difficult in terms of exploration as the goal image does not directly provide the direction to explore. Another difficulty is exploring the `stop' action. Prior setups of the Image Goal task where end-to-end RL policies were shown to perform reasonably well assumed that the agent succeeded if it hits the goal state. However, based on the suggestion from ~\cite{anderson2018evaluation}, we added the `stop' action as it is more realistic. To quantify the effect of stop action, we report the performance of all the models without the stop action in Table~\ref{tab:results_stop}(left). We see that the performance of RL baselines is much higher. However, the performance of \methodabbr{} also increases as the agent automatically stops when it reaches the goal state instead of using the prediction of the Relative Pose Estimator to stop. Other differences that make our experimental setup more realistic but also makes exploration harder for RL as compared to prior setups include continuous state space as compared to grid-based state space, fine-grained action as compared to 90 degree turns and grid cell forward steps and stochastic transitions due to realistic motion noise.\\\vspace{-6pt}

\noindent \textbf{Comparison with spatial map-based methods and the effect of motion noise.} The performance of metric spatial map-based baselines drops quickly as the distance to the goal increases. This is likely due to the accumulation of pose errors as the trajectory length increases. Errors in pose prediction make the map noisy and eventually lead to incorrect path planning. To quantify this effect, we evaluate all the models without any motion actuation and sensor noise in Table~\ref{tab:results_stop}(right). The results show that the performance of the metric map-based baselines scales much better with distance in the absence of motion noise, however, the performance of NTS does not increase much. This indicates that \methodabbr{} is able to tackle motion noise relatively well. This is because \methodabbr{} only uses pose estimates between consecutive nodes, which do not accumulate much noise as they are a few actions away from each other. The performance of \methodabbr{} is still better than the baselines even without motion noise as it consists of Semantic Score Predictor ($\mathcal{F}_S$) capable of learning structural priors which the metric spatial map-based baselines lack. We quantify the effect of the Semantic Score Predictor in the following subsection.

\subsection{Ablations and Sequential Goals}
\vspace{-4pt}
\noindent In this subsection, we evaluate the proposed model on sequential goals in a single episode and study the importance of the topological map or the graph and the Semantic Score Predictor ($\mathcal{F}_S$). For creating a test episode with sequential goals, we randomly sample a goal between $1.5m$ to $5m$ away from the last goal. The agent gets a time budget of 500 timesteps for each goal. We consider two ablations:\\\vspace{-9pt}

\noindent \textbf{\methodabbr{} w/o Graph.} We pick the direction with the highest score in the current image greedily, not updating or using the graph over time. Intuitively, the performance of this ablation should deteriorate as the number of sequential goals increases as it has no memory of past observations.\\\vspace{-9pt}

\noindent \textbf{\methodname{} w/o Score Function.} In this ablation, we do not use the Semantic Score Predictor ($\mathcal{F}_S$) and pick a ghost node randomly as the long-term goal when the Goal Image is not localized in the current graph. Intuitively, the performance of this ablation should improve with the increase in the number of sequential goals, as random exploration would build the graph over time and increase the likelihood of the Goal Image being localized.\\\vspace{-9pt}

\begin{figure*}[h!]
\centering
\includegraphics[width=\linewidth,height=\textheight,keepaspectratio]{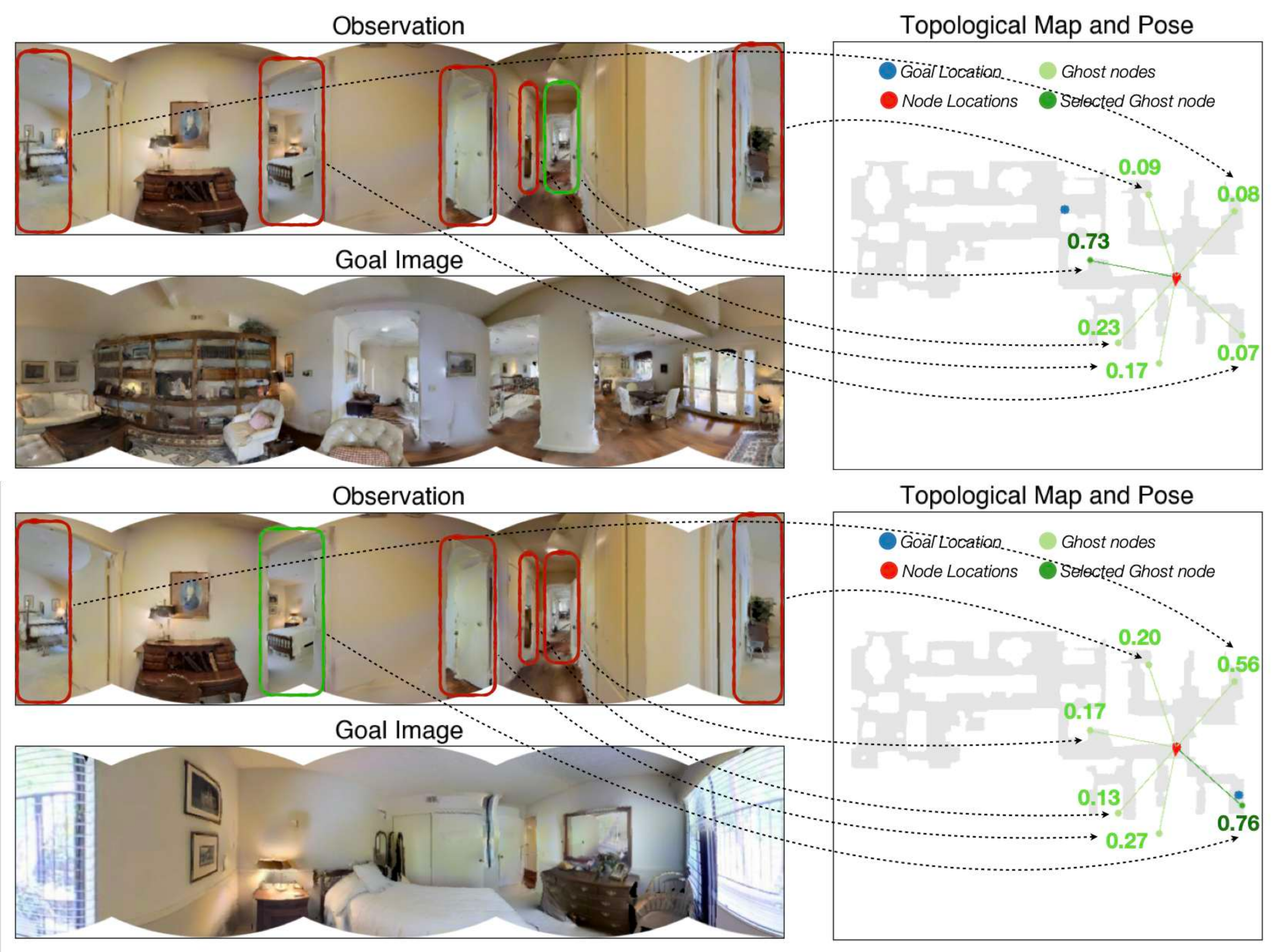}
	\vspace{-12pt}
	\caption{\small \textbf{Learning Semantic Priors.} Figure showing an example indicating that NTS model learns semantic priors. The figure shows the first time step for two different episodes starting at the same location but with different goal images. \textbf{Top:} The goal image is of a living room. The NTS model successfully selects the ghost node leading down the hallway as other ghost nodes lead to different bedrooms. \textbf{Bottom:} The goal images is of a bedroom. With the same starting location and the same ghost nodes, the NTS models select a different ghost node which leads to the correct bedroom.
    }
\label{fig:semantic_priors}
\end{figure*}

We report the success rate and SPL of \methodabbr{} and the two ablations as a function of the number of sequential goals in Figure~\ref{fig:ablation}. Success, in this case, is defined as the ratio of goals reached by the agent across a test set of 1000 episodes. The performance of \methodabbr{} is considerably higher than both the ablations, indicating the importance of both the components. The drop in performance when the semantic score function is removed indicates that the model learns some semantic priors (Figure~\ref{fig:semantic_priors} shows an example). The performance of all the models decreases with an increase in the number of sequential goals because if the agent fails to reach an intermediate goal, there is a high chance that the subsequent goals are farther away. The performance gap between \methodabbr{} and \methodabbr{} w/o Score Function decreases and the performance gap between \methodabbr{} and \methodabbr{} w/o Graph increases with increase in the number of sequential goals as expected. This indicates that the topological map becomes more important over time as the agent explores a new environment, and while the Semantic Score Predictor is the most important at the beginning to explore efficiently.

\vspace{-2pt}
\section{Discussion}
\vspace{-4pt}
\noindent We designed topological representations for space that leverage semantics and afford coarse geometric reasoning. We showed how we can build such representation autonomously and use them for the task of image-goal navigation. Topological representations provided robustness to actuation noise, while semantic features stored at nodes allowed the use of statistical regularities for efficient exploration in novel environments. We showed how advances made in this paper make it possible to study this task in settings where no prior experience from the environment is available, resulting in a relative improvement of over $50\%$. In the future, we plan to deploy our models on real robots.

\vspace{-2pt}
\section*{Acknowledgements}
\vspace{-4pt}
\noindent This work was supported  in part by IARPA DIVA D17PC00340, US Army
W911NF1920104, ONR Grant N000141812861, ONR MURI, ONR Young Investigator, DARPA MCS, Apple and Nvidia.

\noindent\textbf{Gibson License}: {\footnotesize \url{http://svl.stanford.edu/gibson2/assets/GDS_agreement.pdf}}
\newpage
{\small
\bibliographystyle{ieee_fullname}
\bibliography{references}
}

\clearpage
\appendix

\section{NTS Multi-task Learning Model Architecture and Training Details}
\begin{figure*}[h!]
\centering
\includegraphics[width=\linewidth,height=\textheight,keepaspectratio]{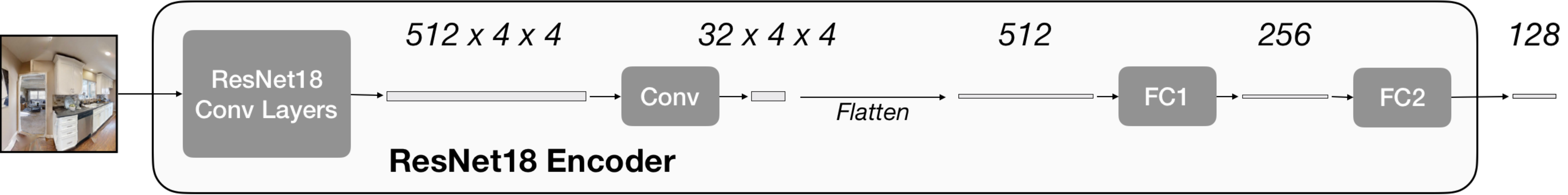}
\caption{\small \textbf{ResNet18 Encoder.}  Figure showing the architecture of the ResNet18 Encoder.}
\label{fig:resnet18_encoder}
	\vspace{-10pt}
\end{figure*}

\vspace{-4pt}
The NTS Multi-task Learning model consists of 4 modules, a ResNet18 encoder~\cite{he2016deep}, a Connection model, an Inter-node Predictions model, and an Intra-node Predictions model as shown in Figure~\ref{fig:model}. Given panoramic source ($I_S$) and goal ($I_G$) images, each of size $128 \times 512$, we first construct $n_\theta = 12$ square patches from each panoramic image. The $i^{th}$ patch consists of $128 \times 128$ image centered at $i / n_{\theta} * 2\pi $ radians. These resulting patches are such that each patch overlaps with 1/3 portion of the adjacent patch on both sides (patches are wrapped around at the edge of the panorama). Each of the $n_\theta$ patches of both source and target images is passed through the ResNet18 encoder to get a patch representation of size 128. The architecture of the ResNet18 encoder is shown in Figure~\ref{fig:resnet18_encoder} for RGB setting. For RGBD, we pass the depth channel through separate non-pretrained ResNet18 Conv layers, followed by 1x1 convolution to get a representation of size 512. This representation is concatenated with the 512 size RGB representation and passed through FC1 and FC2 to finally get the same 128 size patch representation. We use a dropout of 0.5 in FC1 and FC2 and ReLU non-linearity between all layers.

\begin{figure*}[h!]
\includegraphics[height=5cm,keepaspectratio]{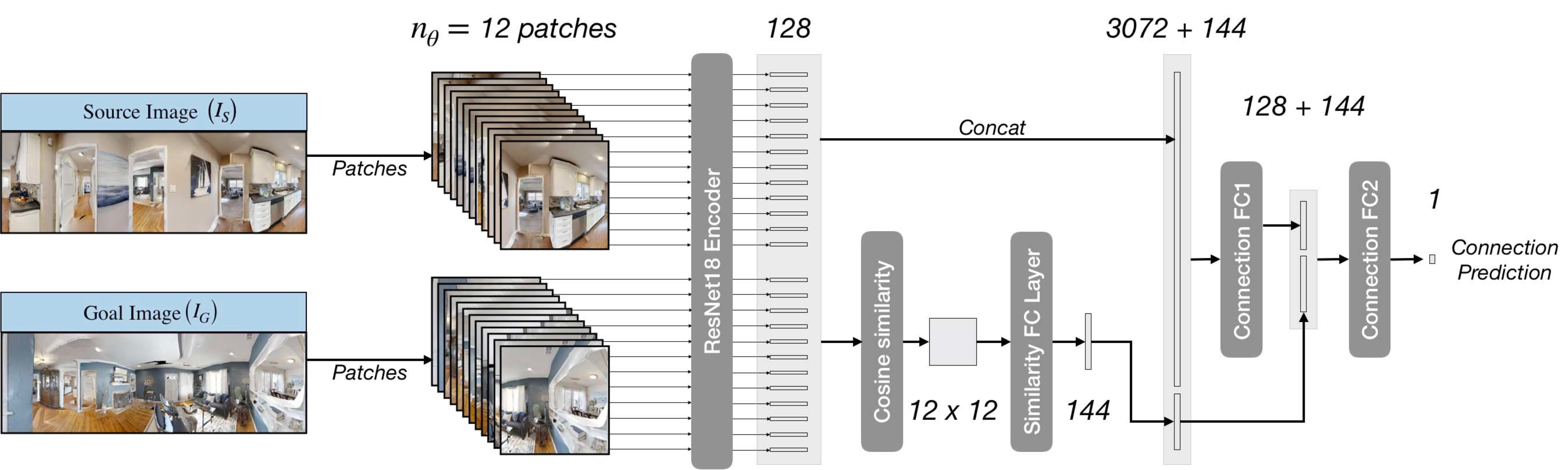}
\vspace{3pt}
\caption{\small \textbf{Connection Model.} Figure showing the architecture of the Connection Model.}
\label{fig:connection}
	\vspace{-10pt}
\end{figure*}

\begin{figure*}[h!]
\includegraphics[height=5cm,keepaspectratio]{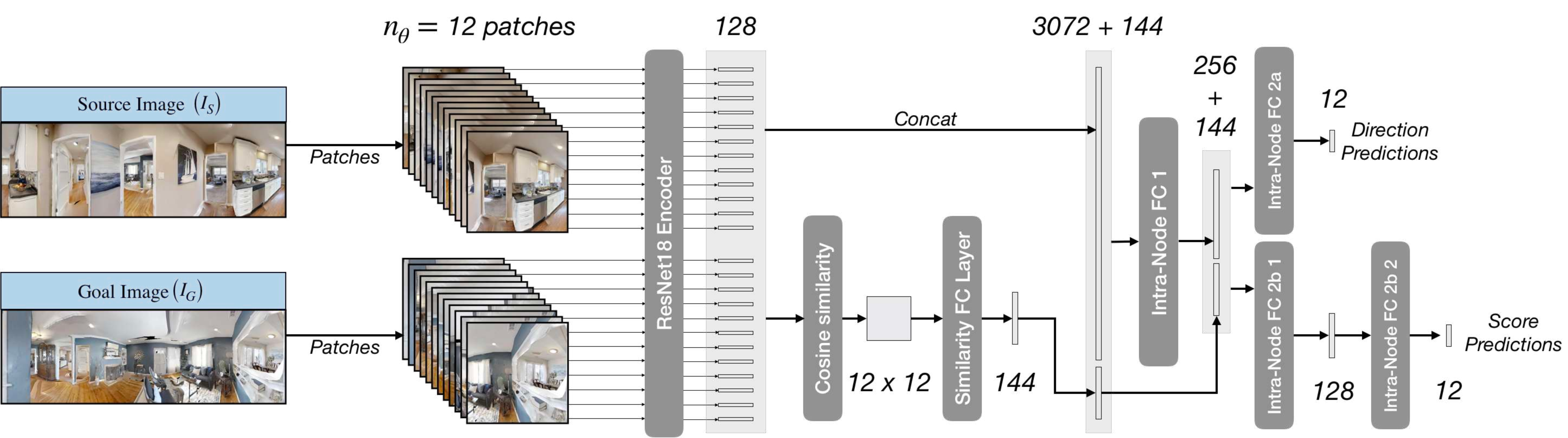}
\vspace{3pt}
\caption{\small \textbf{Intra-Node Predictions Model.} Figure showing the architecture of the Intra-Node Predictions Model.}
\label{fig:intra_node}
	\vspace{-10pt}
\end{figure*}

\begin{figure*}[h!]
\includegraphics[height=5.46cm,keepaspectratio]{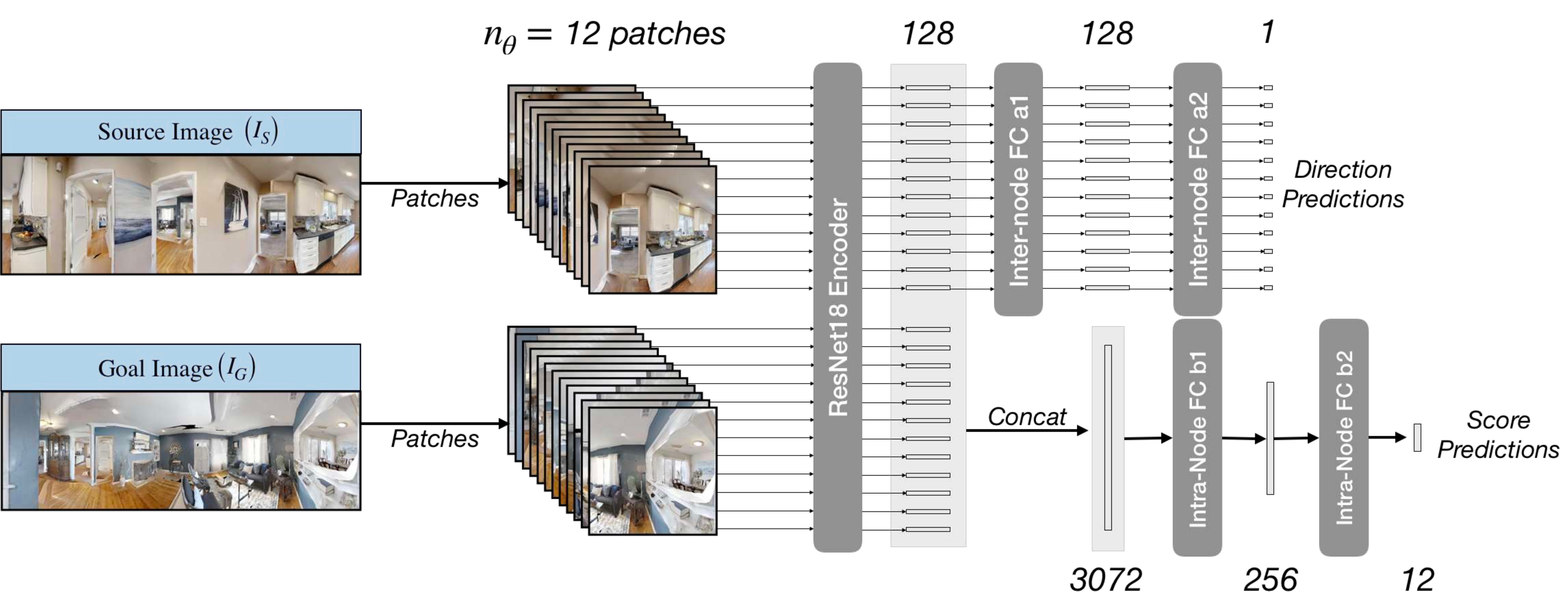}
\caption{\small \textbf{Inter-Node Predictions Model.} Figure showing the architecture of the Inter-Node Predictions Model.}
\label{fig:inter_node}
	\vspace{-5pt}
\end{figure*}

Given 12 patch representations for source and goal images each, the connection model predicts whether the two images belong to the same node or not. The architecture of the Connection Model is shown in Figure~\ref{fig:connection}. If the two images belong to the same node, the Intra-Node Predictions model predicts the direction and score (or equivalently distance) of the goal image relative to the source image. The architecture of the Intra-Node Predictions Model is shown in Figure~\ref{fig:intra_node}. If the two images belong to different nodes, the Inter-Node Predictions model predicts the explorable area directions and the score of each direction. The architecture of the Inter-Node Predictions Model is shown in Figure~\ref{fig:inter_node}. As shown in the figure, the explorable area directions are specific to the source image patches and independent of the goal image. We use ReLU non-linearity between all layers in all the modules.

The entire model is trained using supervised learning. We use Cross-Entropy Loss for the Connection and direction labels and MSE Loss for score labels. We use a loss coefficient of 10 for the score labels and a loss coefficient of 1 for the connection and direction labels. We train the model jointly with all the 5 losses using the Adam optimizer~\cite{kingma2014adam} with a learning rate of 5e-4 and a batch size of 64.

\vspace{-2pt}
\section{Dataset collection and automated labeling}
\vspace{-4pt}
\noindent For training the NTS Multi-task Learning model, we need to collect the data and different types of labels. We randomly sample 300 points in each training scene. For each ordered pair of images, source image ($I_S$) and goal image ($I_G$), we gather the different labels as follow:\\\vspace{-9pt}

\noindent\textbf{Connection label:} This label denotes whether the source and goal image belong to the same node or not. We get this label by detecting whether the goal image location is visible from the source image. To compute this, we take a 5-degree patch of depth image centered at the relative direction of the goal image from the source image. If the maximum depth in this patch is greater than the distance to the goal image, then the goal image belongs to the same node as the source image and the label is 1. Intuitively, the above checks whether the maximum depth value in the direction of the goal image is greater than the distance to the goal. If the maximum depth in the direction of goal image is greater than the distance to the goal image or if the distance to the goal image is greater than $r=3m$, then the goal image belongs to a different node and the label is 0.\\\vspace{-9pt}

\noindent\textbf{Intra-node labels:} If the source and goal image belong to the same node, i.e. the above label is 1, then the intra-node direction and score labels are directly obtained from relative position of the goal image from the source image. Let the relative position of the goal image be $\Delta p = (d, \theta)$, where $d$ is the distance and $\theta$ is the angle to the goal image from the source image. Intra-node direction label is just the $360/n_{\theta} = 30$ degree bin in which $\theta$ falls, i.e. $nint(\theta/2\pi \times n_{\theta})$ where $nint(\cdot)$ is the nearest integer function. For the intra-node score label ($s$), we just convert the distance $d$ to a score between $0$ and $1$ using the following function:
$$ s = max((1 - d/r), 0)$$
\noindent where $s$ denotes the score, $r = 3m$ denotes the radius of the node, $d$ is the distance. Note that the direction label is discrete and the score label is continuous. \\\vspace{-9pt}

\noindent\textbf{Inter-node labels:} If the source and goal image do not belong to the same node, i.e. the connection label is 0, we compute the inter-node labels. For computing the inter-node direction labels, we first project the depth channel in the panoramic source image to compute an obstacle map. We ignore obstacles beyond $r=3m$. In this obstacle map, we take $n_{\theta} = 12$ points at angles $i/n_{\theta} \times 2\pi, \forall i \in [1,2,\ldots,12]$ and distance $r=3m$ away. For every, $i \in [1,2,\ldots,12]$, if the shortest path distance to the corresponding is less than $1.05\times r$, then the corresponding inter-node direction label is 1 and otherwise 0.

Let the inter-score labels be denoted by $s_i, \forall i \in [1,2,\ldots,12]$. For the inter-node score label $s_i$, we find the farthest explored and traversable point in direction $i \times (2\pi/n_{\theta})$ in the above obstacle map. Let the shortest path distance (geodesic distance) of the goal image from this point be $d_i$. Then the score label is given by:
$$ s_i = max((1 - d_i/d_{max}), 0)$$
\noindent where $d_{max} = 20m$ is the maximum distance above which the score is 0.

\end{document}